# COMPARISON OF OBJECT DETECTION METHODS FOR CORN DAMAGE ASSESSMENT USING DEEP LEARNING

Ali Hamidisepehr[1,2], Seyed Vahid Mirnezami[2,3], Jason K. Ward[4]

[1]Agricultural Engineering Department, Postdoctoral Scholar, North Carolina State University, Raleigh, NC, USA, ali.hamidisepehr@gmail.com
[2]Colaberry Inc, Boston, MA, USA
[3]Mechanical Engineering Department, Iowa State University, Ames, IA, USA, vahid.gvg@gmail.com
[4]Agricultural Engineering Department, Assistant Professor, North Carolina State University, Raleigh, NC, USA, jason.ward@ncsu.edu

**HIGHLIGHTS**
- Corn damage detection using advanced deep learning and computer vision techniques customized for detecting hurricane and flooding damage in farms
- Over 95% accuracy using YOLO model
- Providing useful information to farmers and insurance companies for getting an accurate insight of the corn farm using UAS imagery

**ABSTRACT.**

*Severe weather events can cause large financial losses to farmers. Detailed information on the location and severity of damage will assist farmers, insurance companies, and disaster response agencies in making wise post-damage decisions. The goal of this study was a proof-of-concept to detect damaged corn areas from aerial imagery using computer vision and deep learning techniques. A specific objective is to compare existing object detection algorithms to determine which is best suited for corn damage detection. Two modes of crop damage common in maize (corn) production were simulated: stalk lodging at the lowest ear and stalk lodging at ground level. Simulated damage was used to create a training and analysis data set. An unmanned aerial system (UAS) equipped with an RGB camera was used for image acquisition. Three popular object detectors (Faster R-CNN, YOLOv2, and RetinaNet) were assessed for their ability to detect damaged regions in a field. Average precision was used to compare object detectors. YOLOv2 and RetinaNet were able to detect corn damage across multiple late-season growth stages. Faster R-CNN was not successful unlike the other two detectors. Detecting corn damage at later growth stages was more difficult for all tested object detectors. Weed pressure in simulated damage plots and increased target density added additional complexity.*

*Keywords.*

**INTRODUCTION**

Digital farming techniques are becoming increasingly useful as advances in sensing, data processing, and analytics are becoming more accessible. Crop producers whose operations are exposed to severe weather can use data-driven approaches to assess the impact of acute crop damage events and to respond faster than manual detection of crop damage with more information. The need for digital agriculture tools after severe weather events was painfully evident to producers along the U.S. Atlantic seaboard who were exposed to multiple severe hurricane seasons. Tropical weather events in the U.S. coastal states like North Carolina most often occur during harvest season (Stewart, 2017; Stewart & Berg, 2019) (Figure 1a). The 2018 hurricanes Florence and Michael inflicted over a billion dollars in losses to North Carolina's agricultural industry. Even when acute weather events are less impactful, wildlife can damage crops and create financial losses late in the growing season when, for example, black bears feed on corn to add fat prior to winter (Figure 1b). New tools are needed which can detect and quantify crop damage late in the growing season near to harvest.

Current methods for detecting and reporting crop damage are manual and visual. After a severe weather event growers report that damage has occurred and then different stakeholders such as insurance adjusters, extension agents, or disaster response agencies will survey the field by walking through or driving along the damaged field. Geospatial information such as field boundaries, damaged area boundaries, or geo-tagged descriptive images may be collected.

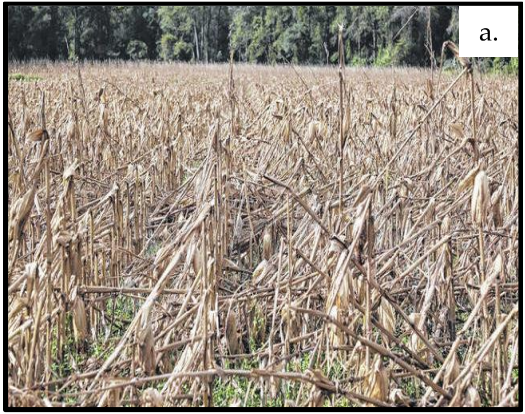 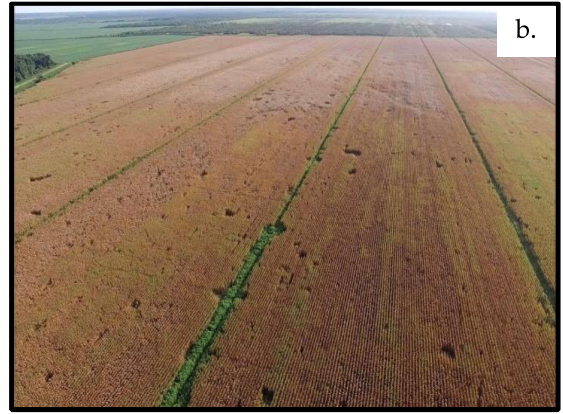

Figure 1: Examples of crop damage caused by: a. Hurricane Florence (Source: Sherry Matthews, The Sampson County Independent, 19 Sept., 2018); b. Black bears flatten corn stalks while feeding on corn in Eastern North Carolina (NC Blackbear Newsletter, 31 Aug., 2018).

Detailed information on the presence and severity of crop damage would help producers to make the decision to harvest the crop or to make an insurance claim on the field. Insurance providers and emergency response agencies can use crop damage information to estimate payouts or to report damage faster which can increase the effectiveness of the support for severe weather impacted communities. Manned aircraft and satellites have been used for collecting remotely sensed data but are limited by cost and spatial and temporal resolution (Hamidisepehr et al., 2017). Unmanned aerial systems (UASs) have become a popular remote sensing tool in digital agriculture which enable growers to have precise information about their fields at specific times of interest (Hamidisepehr & Sama, 2018a). There are a wide variety of sensors which can be deployed to measure different field parameters such as hyperspectral, multispectral, and thermal imagery (Hamidisepehr & Sama, 2018b). The standard digital camera set-up, consisting of red, green, and blue (RGB) bands is the most recognized data to end-users and provide images in the human-visible range at an affordable price. When combined with a UAS platform, an RGB payload can provides a high spatial resolution survey of a field when many individual images with a high overlap are stitched together (Mahajan et al., 2015).

Traditional image processing methods utilized manually extracted target data and static methods for analysis (Ma et al., 2019; Vibhute & Bodhe, 2012; Y. Zhou et al., 2019). As dataset size and analysis complexity increased, traditional image processing methods were less effective or they failed in robustly processing large datasets and complex images (Kamilaris & Prenafeta-Boldú, 2018). Computer vision has significantly improved the power of image processing tools. Computer vision utilizes algorithms to address various tasks such as image detection (Jayas et al., 2000), segmentation (Sammouda et al., 2014), and classification (Krizhevsky et al., 2012). There are several examples using computer vision in digital agriculture including weed detection (Lu et al., 2017), disease detection (Mohanty et al., 2016), plant recognition (Reyes et al., 2015), plant segmentation (Falk et al., 2020), fruit counting (Rahnemoonfar & Sheppard, 2017), and crop classification (Rebetez et al., 2016).

Object detection is a computer vision technique that is applied for detecting specific objects in images. Face detection (Kazemi & Sullivan, 2014) and pedestrian detection (Li et al., 2016) are the most well-developed applications of this technique. The most common object detection methods include Faster Region-based Convolutional Neural Network (Faster R-CNN) (Ren et al., 2015), You Only Look Once Version 2 (YOLOv2) (Redmon & Farhadi, 2017), and RetinaNet (Lin et al., 2017). Each of these methods perform differently under different applications. YOLOv2 and RetinaNet can perform faster than Faster R-CNN because they implemented a single stage detection process. Additionally, they showed higher performance on standard dataset (Kamilaris & Prenafeta-Boldú, 2018; Zheng et al., 2018). However, Faster R-CNN has been a popular method for several applications due to ease of use (Kamilaris & Prenafeta-Boldú, 2018). In most deep learning applications, it is common to utilize a pre-developed computer vision model trained on a relevant dataset (so-called transfer learning). Collecting a large enough dataset for developing a custom deep learning method would be difficult, time-consuming, a nearly impossible for most users focused on application. Using transfer learning, existing feature extraction methods, such as those mentioned previously, can be leveraged from models trained on standard datasets and object detection is fine-tuned to the desired target. (Kamilaris & Prenafeta-Boldú, 2018).

Agricultural applications of object detection are becoming more common. A real time vegetable detection system was developed using deep learning networks (Zheng et al., 2018). Multiple object detectors were selected in order to recognize tomato and cucumber at different stages. Among all advanced detectors selected, YOLOv2 had the highest performance in terms of model average precision (AP). Koirala et al. [23], tested these main detection algorithms to detect mangos in images. By optimizing YOLOv2, they developed a new algorithm which exhibited improved detection performance. RetinaNet was also tested in vineyards to detect Esca disease and obtained AP of 70% (Rançon et al., 2019). AP is an index that incorporates the ability of a model to make correct classifications and the ability to find all relevant objects.

The combination of computer vision, deep learning, and powerful hardware have demonstrated success in digital agriculture projects. Using computer vision techniques along with UAS imagery will allow precise assessment of field conditions (Tripicchio et al., 2015). Nolan [26] used computer vision techniques on a UAS to delineate vine-rows automatically and proved the efficiency of the system in commercial vineyards. An autonomous UAS with onboard computer vision capabilities was recently developed. This system provided promising results for weed detection and color detection for fruit sorting; however, authors highly recommended using machine learning methods instead of the traditional image processing method (Target Detection Software) for a more robust weed detection system (Alsalam et al., 2017).

Identifying crop damage caused by severe weather conditions or other stressors via remote sensing has been also tested. For instance, the structure-from-motion photogrammetry method was applied to detect lodging in maize. In this approach a 3D map needs to be created after stitching individual and overlapping images, collected from RGB and NIR cameras (Chu, Starek, Brewer, Masiane, et al., 2017). Multispectral imaging was used to detect crop hail damage using vegetation indices. Detection was more precise in cases with more severe damage or when imagery was acquired the early days after damage was occurred (J. Zhou et al., 2016). Crop damage was assessed using satellite imagery after a frost but low spatial and temporal resolution made future applications of the technique unlikely (Silleos et al., 2002).

Despite the progress in recent years, crop damage identification has room for improvement. It can be a place for deep learning and object detection methods to be deployed for detecting and analyzing crop damage data at different stages of growth and with more complexity. In this study, the ability for object detection methods for corn damage detection in late-season corn at different stages of senescence was analyzed. Individual methods were compared in terms of their prediction power.

The specific objectives were:

1. To compare object detection models performance for damage measurement in the corn field,
2. To assess the best model and best growth stage for damage detection in the corn field.

## MATERIALS AND METHODS

### STYLES FIELD LOCATION AND PLOT LAYOUT

The field study site was located near Goldsboro, NC, US, on the Cherry Research Station. Corn hybrid Augusta 5065, 115 day relative maturity (RM), was planted on 19 April, 2018. Seed population was 80,000 seeds/ha (32,000 seeds per acre) on 76.2 cm (30 in.) row spacings. Plots were established in strips six rows wide and 46 m (150 ft) long for a total area of ~210 m$^2$ per treatment strip. The treatment areas were rectangular. The shape of the simulated treatment area has no bearing on the training and detection of damaged regions within the field. Even if the simulated lodging region was a complex shape, the training annotation and object detection bounding boxes were rectangular and parallel (or near parallel) to image boundaries. Treatments were not replicated as the goal of this study was to initially determine if physical corn damage could be reliably detected in imagery and to compare object detection methods.

### FIELD DAMAGE SIMULATION

Eleven treatments were created to simulate two different physical corn damage modes at five different late-season growth stages, along with a control. Two corn damage modes were simulated to represent different weather impacts on a corn crop lodging near harvest. Standing stalks were broken immediately under the first ear or at ground level. Simulated damage was created by manually breaking every corn stalk in the plot in the mode designated by the treatment plan. Time effects were referenced to crop physical maturity as estimated by hybrid RM: stage one was two weeks prior to the week of RM, stage two was one week prior, stage three was the week of RM, stage four was the week after RM, and stage five was two weeks post RM. Treatments were chosen to add diversity to the object detection model from physical crop differences and the impact of crop senescence over time. There was only one treatment by time combination because this project was an initial assessment to determine if physical corn damage could be reliably detected in imagery and to compare object detection methods.

**IMAGE ACQUISITION**

UAS (M600 Pro, DJI, Shenzen, China) imagery was collected using an RGB camera (Zenmuse X5, DJI, Shenzen, China). Imagery was collected at 92 m above ground level at a ground sampling distance of 2.25 cm/px (2). Imagery was collected on the same day immediately after that growth stage's treatments were applied (Table 1). Figure 2 shows two examples showing the farm at different at different stages. One example when the camera angle is straight, and the other example is taken from a turning position.

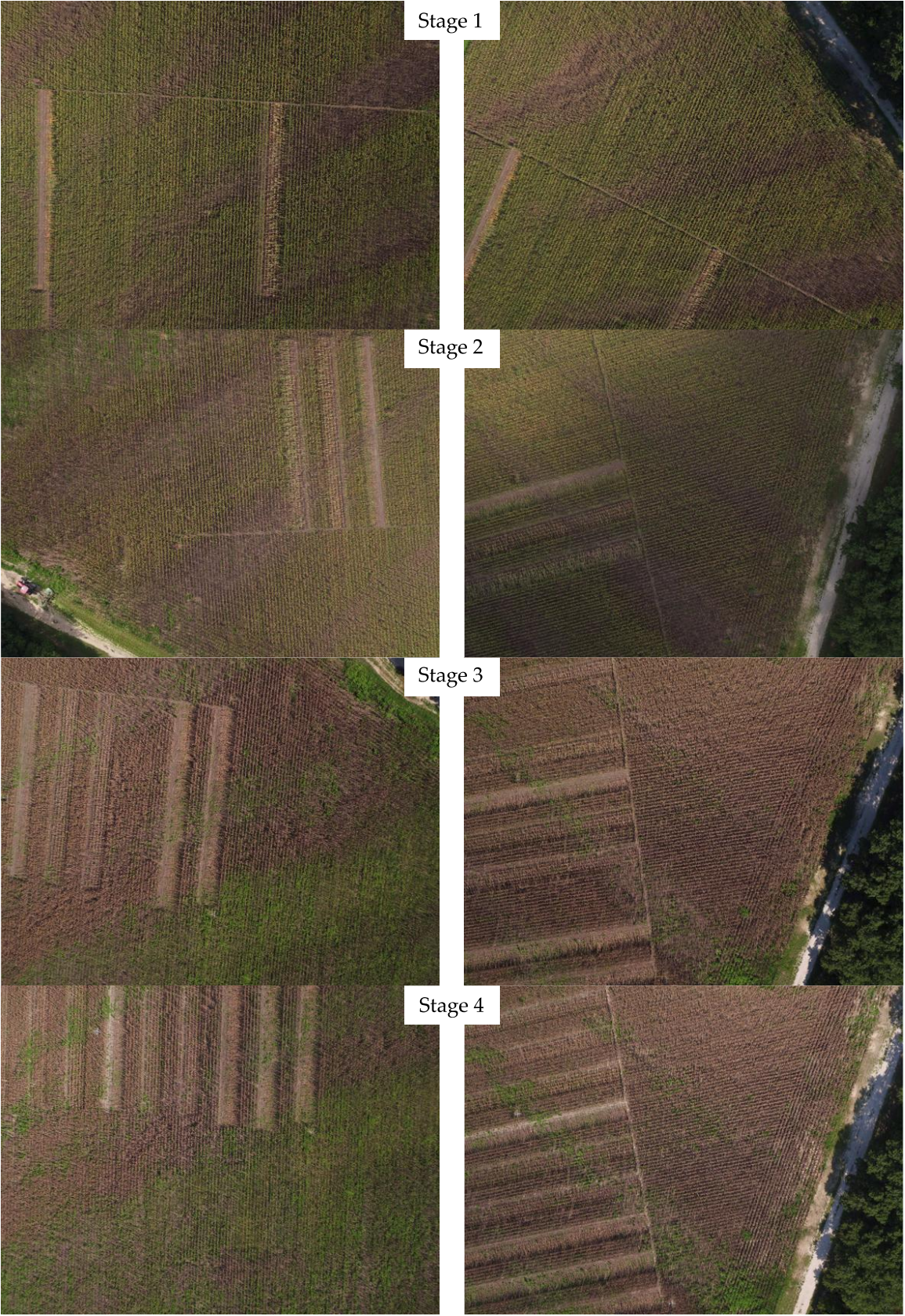

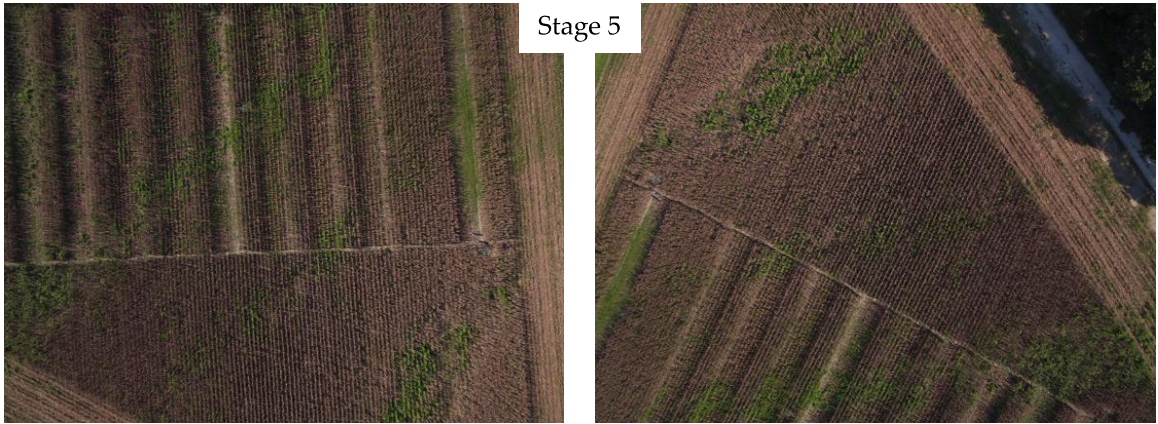

**Figure 2. Examples of aerial imagery at different stages of the growing season**

**Table 1. Date and days after planting (DAP) of treatment and data collection events during the 2018 growing season.**

| Activity | Date | DAP |
|---|---|---|
| Planting | 19 Apr | 0 |
| Stage 1 | 3 Aug | 106 |
| Stage 2 | 10 Aug | 113 |
| Stage 3 | 17 Aug | 120 |
| Stage 4 | 24 Aug | 127 |
| Stage 5 | 31 Aug | 134 |

**DATA PREPROCESSING**

Images which did not contain a damaged region (known as negative images) or were collected before reaching the desired altitude were excluded. This was done because the number of negative images were substantially higher than the images with damaged regions in them (known as positive images). Even in images with damaged regions the majority of the image area was negative for target objects. Negative images would not enhance object detection model performance and would cause longer training time. The goal was to assess object detection methods, not detection of the regions without target objects. False positive results could be adequately assessed from the images containing target objects.

Model refers to a successfully trained object detector using one of the three architectures at a particular stage. The YOLOv2 Stage 1 model was an object detector built using YOLOv2 trained using annotated images from the Stage 1 data collection event only. An additional model for each architecture was developed based on all training images from all growth stages to determine if it would provide greater prediction power because of increased diversity in imagery data.

A limitation with the tested object detectors were that input images could only be annotated for training or identify target objects with rectangular areas whose borders were parallel (or near parallel) to the image borders. A complex shape was still identified with a rectangle during training and testing. Object orientation and dimensions, however, could create complexity. A region whose major axis was not parallel to the image borders increased the area which was incorrectly labeled as the target object. Negative space outside of the target regions but within the label area got included in the object classification. Damaged areas were rectangular and UAS flight path was almost parallel to the major axis of rectangular damaged area. Although during each flight, multiple images were taken at the end of each flight line while the UAS was turning. To analyze the impact of these images in model performance, an additional model for each stage was created in which these images were removed. Considering all stages including 5 models for each filtered and unfiltered datasets plus aggregated models (models trained by images from all stages together), 12 models were generated for each object detector. Filtering refers to removing non-parallel images from each dataset. Images taken in a turning position were the difference between filtered and unfiltered datasets. It was expected that images in an turning angle might affect the performance of models due to different looks. The number of images in each model without filtering can be seen in Table 2. Creating simulated damaged regions for research purposes and then collecting imagery data is very labor intensive. Thus, in order to make maximum use of this data, new data was derived out of original data using data augmentation techniques. New images were generated randomly during the training process by rotation, cropping, changing color, and resizing of original images. Data augmentation was provided as a built-in feature and was used to compensate for having limited dataset. The data augmentation hyperparameters were left by their default values.

Original images were resized to 12.5% of their original size (from 4600×3400 px to 570×430 px) to make the training process faster and hardware usage more efficient.

Table 2. The number of images and bounding boxes at different stages without filtering

| | Number of images | Number of bounding boxes |
| --- | --- | --- |
| Stage 1 | 80 | 112 |
| Stage 2 | 92 | 267 |
| Stage 3 | 103 | 540 |
| Stage 4 | 103 | 681 |
| Stage 5 | 100 | 670 |
| Total | 478 | 2270 |

## DATA ANNOTATION

In order to prepare image datasets for training, damaged areas in individual images were labeled by rectangular bounding boxes. The different object detectors required different formats for labeled images. For Faster R-CNN models, images were labeled using Image Labeler, a MATLAB application. A MATLAB script was used to convert Faster R-CNN labels into the format required by RetinaNet. YOLOv2 required a distinct format, therefore a different image labeling tool was needed (LabelImg). Figure 3 shows an overview of the corn damage detection system starting with data collection and annotation. Feature extractors were chosen based on the detector method. Resnet50, VGG16, and Darknet19 were chosen for Faster RCNN, RetinaNet, and YOLO, respectively similar to Zheng et al. (2018).

## TRAINING AND TESTING

The dataset for each model was randomly divided to use 85% of the images for training and 15% for testing as in Hamidisepehr (2018). Training and testing datasets were randomly selected for 3 times and the final performance was obtained by taking the average from running 3 different data selection. Hyperparameters were variables needed to fine-tune weights from pre-developed models like Resnet and VGG before applying a learning algorithm to a custom dataset. Hyperparameters including number of epochs, batch size, and learning rate were adjusted on each training dataset to deliver the optimal solution at each stage. Initial values were set for each model based on default values suggested by Lin et al. (2017).

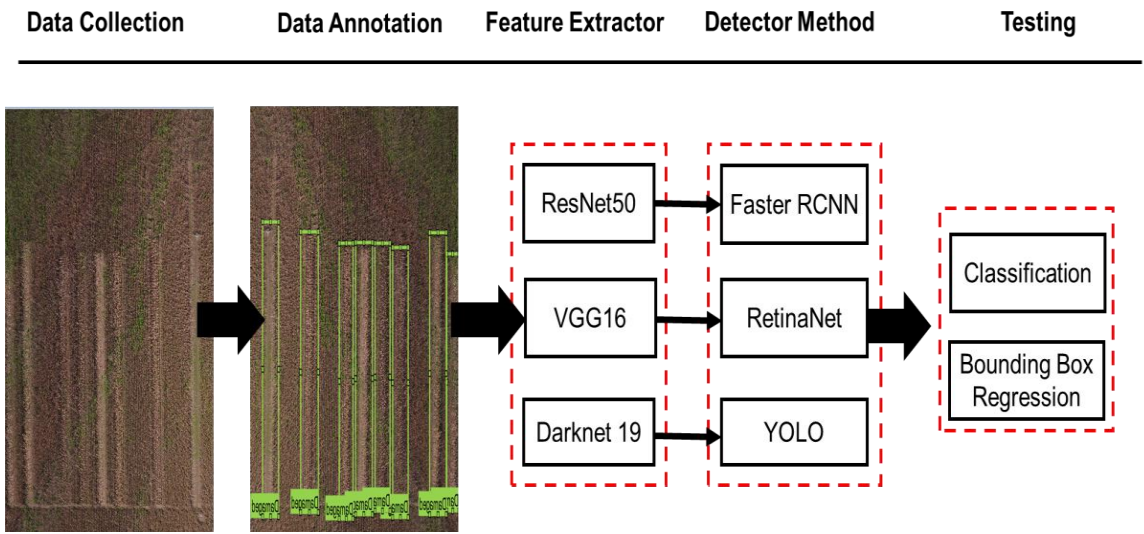

Figure 3. The structure flow of the corn damage recognition system based on deep learning and computer vision. The architecture of the system includes input image, damage annotation, feature extraction, training, and detection of corn damage in each image. B-box is bounding box in a short form. All bounding boxes are labeled as "damaged" in annotations section.

Epochs represented the number of times that the entire training datasets passed through a neural network. Increasing the number of epochs results in greater prediction power, however, an excessively large number of epochs increased the training time with no performance improvement (Amiri et al., 2017). Model performance was quantified by calculating total loss. Total loss combined the error in bounding box location, size, and confidence which was the probability of existence of the desired object in the bounding box. The amount of loss was monitored after each epoch and training ceased if the total loss did not marginally decrease. Batch size described the number of images presented to the learning algorithm in one pass. Increasing the batch size usually resulted in higher prediction power; although, extreme batch size would cause memory error. Small batch size, which is needed on less powerful hardware, introduced undesirable noise which prevented the training process from converging to an optimal value (Rhu et al., 2016). Learning rate specified the rate of training. Similar to batch size, increasing learning rate can improve the model performance but can causes memory errors depending on hardware capacity (Dauphin et al., 2015).

Number of epochs, batch size, and learning rate were determined empirically to obtain an optimal solution with high performance and minimal training time without memory error. Table 3 shows the hyperparameters set for each object detector.

**Table 3: Hyperparameters setting based on preliminary tests**

| Model | Epochs | Batch Size | Learning Rate |
|---|---|---|---|
| Faster R-CNN | 200 | 1 | $10^{-3}$ |
| YOLOv2 | 200 | 2 | $10^{-5}$ |
| RetinaNet | 200 | 4 | $10^{-5}$ |

## COMPUTING HARDWARE

Training object detection models using deep learning algorithms was computationally intensive and required the use of advanced graphical processing unit (GPU) technology (Hwu & Kirk, 2009). Training was initially attempted using nodes equipped with NVIDIA Tesla P100 and NVIDIA GeForce GTX 1080. These GPUs provided sufficient resources for training Faster R-CNN object detection models but struggled to complete YOLOv2 and RetinaNet models due to larger batch size. Later access was provided to computation nodes with more powerful NVIDIA Tesla V100 GPUs. This hardware was able to train YOLOv2 and RetinaNet object detection models without errors in all models except one YOLO model training on unfiltered images from the entire season. This model did not go through the whole training process even with the smallest batch size due to out of memory error.

## EVALUATION

In order to assess the performance of each model created at different growth stages and with different object detectors, three evaluation metrics were selected: precision, recall, and AP. Precision represents model performance based on the ratio of the number of correct damage detection to the total number of incorrect and correct damage detection. Recall is the ratio of the number of correct damage detections to the total actual damage regions in the field. Precision and Recall are calculated based on prediction parameters including true positive, false positive, and false negative. It is common to prioritize the ultimate goal to either minimize false positive or false negative based on the fault tolerance in a specific project. For example in disease diagnosis minimizing false negative is more important and for spam detection minimizing false positive is critical. F1 score combines precision and recall for a specific class. The F1 score can be interpreted as a weighted average of the precision and recall, where an F1 score reaches its best value at 1 and worst at 0. Equations (1-3) for the performance metrics are below.

$$Precision = \frac{TP}{TP + FP} \quad (1)$$

$$Recall = \frac{TP}{TP + FN} \quad (2)$$

$$F1\ score = 2 \times \frac{Precision \times Recall}{Precision + Recall} \quad (3)$$

where TP (True Positive) was the number of correctly detected corn damage regions, FP (False Positive) was the number of undamaged areas detected as damaged, FN (False Negative) was the number of missed damaged areas. The goal was to maximize TP and minimize FN in a model.

Intersection over union (IoU) measures how much the predicted damaged boundary overlaps with the ground truth. 50% overlap between actual and predicted objects was considered as "match" or a TP object. In this way, the number of TP, FP, and FN, were counted. At large enough sample size, the area under the Precision-Recall curve equals AP. AP provides an index that incorporates the ability of the detector to make correct classifications (precision) and the ability of the detector to find all relevant objects (recall) (Everingham & Winn, 2011; Henderson & Ferrari, 2016; Koenig et al., 2015).

## RESULTS

The aim was to detect the damaged areas from the undamaged background. Different hyperparameter values were tested to most efficiently train each model while avoiding memory errors and obtaining the highest performance. On average, each model took about 3 hours to complete training. YOLOv2 completed training fastest, with RetinaNet and Faster R-CNN being slower which agrees with findings from a previous study (Redmon & Farhadi, 2018).

### PRECISION-RECALL CURVE

Precision and recall were calculated for all predictions in the order of their confidence. A precision-recall curve was plotted for each model. The precision-recall curve demonstrated the tradeoff between precision and recall for varying detection thresholds. Maintaining precision at 1.0 or increasing precision indicated that predictions were correct. Recall values of 1.0 at the endpoint of the curve indicated that all objects were detected and there were no missed objects. An ideal model returns accurate predictions (high precision), as well as detecting all positive objects (high recall).

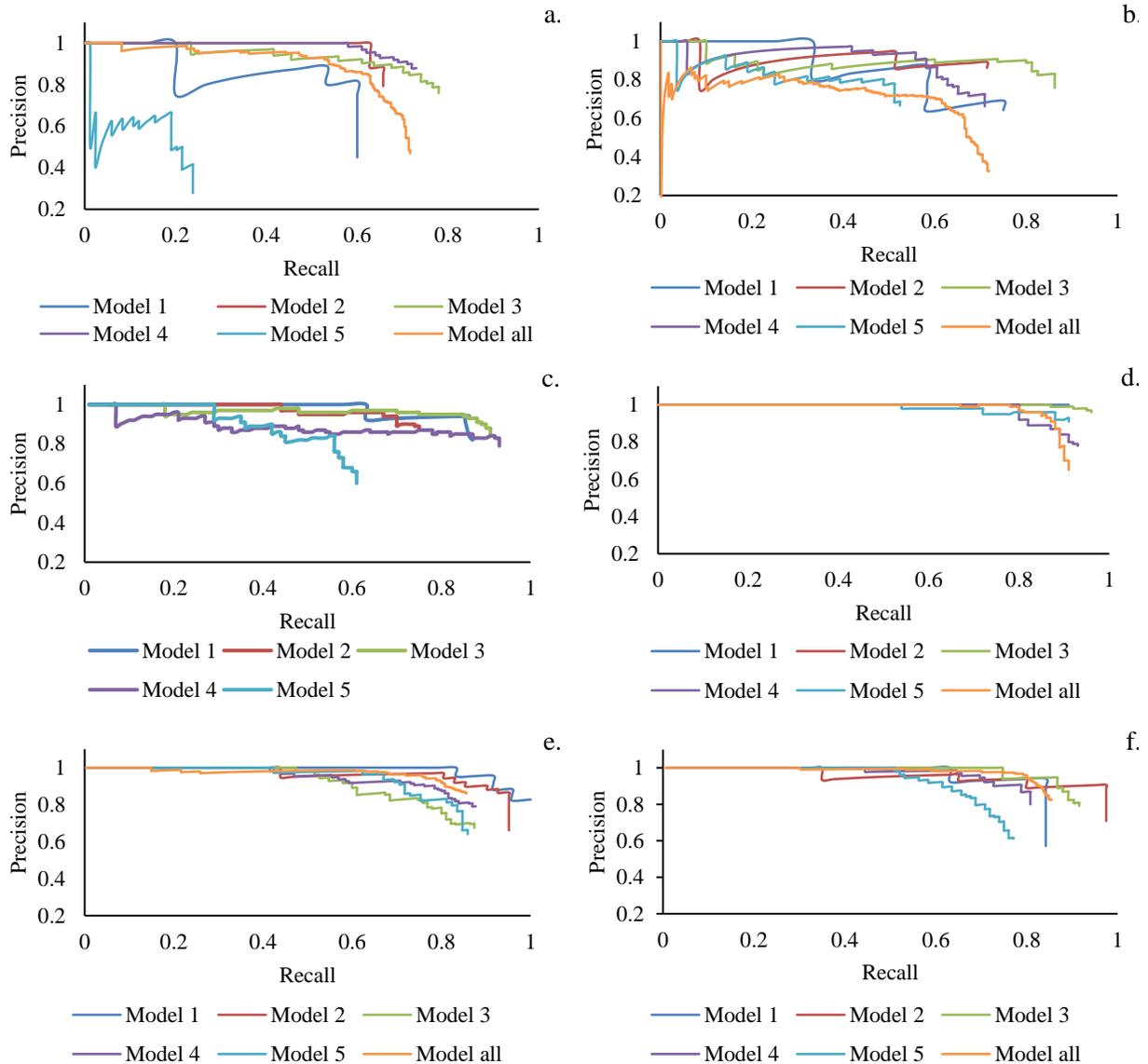

**Figure 4. Precision-Recall curve at different stages for: a. unfiltered Faster R-CNN; b. filtered Faster R-CNN; c. unfiltered YOLOv2; d. filtered YOLOv2; e. unfiltered RetinaNet; f. filtered RetinaNet.**

Figure 4 shows precision-recall curves for each model. In order to obtain a precision-recall curve for each model, all predictions were ranked based on the confidence level the model had for detecting that object regardless if it was a true or false prediction. If a prediction was incorrect the precision value decreased while precision increased for correct decisions. In most cases, objects with top ranks were correct predictions which is why all curves started at 1.0 precision. Recall increased after each correct prediction and remained constant in the case of incorrect predictions. Ideally, precision remained high for all recall values and the curve reached 1.0 recall. In Faster R-CNN models recall values did not reach 1.0 across all growth stages. At least 20% of corn damage regions were not detected. Maximum precision was not maintained for all models, both filtered (Figure 4a, c, and e) and unfiltered (Figure 4b, d, and f) which represented incorrect predictions even with relatively high confidence. YOLOv2, especially after filtering, and RetinaNet remained at high precision at different recall values at different growth stages. Among all stages, stage 5 had the most inaccuracies across different object detectors because of the amount of complexity in the dataset due to number of objects, temporal effect, and closeness of the objects. Other stages provided more accurate models with a slight difference in the overall performance.

Ground truth and predicted corn damage bounding boxes are shown on the same image for visual inspection (Figure 5). Ground truth bounding boxes were the original annotated testing regions. Colors identifying the ground truth and detected damage regions were different across the three algorithms since different labeling and visualization tools were used. Color descriptions are included in the caption. Images include one sample from the testing dataset in stage 3 as an example. The numbers above the boxes in Faster R-CNN and RetinaNet models displayed the confidence level for the detection. YOLOv2 provided this information as well but it was not shown on images. RetinaNet and YOLOv2, Figure 5a and 5b, respectively, detected all damage regions in the image. Figure 5a demonstrated the impact of regions which are not parallel to image borders. The farthest left annotated and detected region crosses two treatment zones. Even with annotated regions including negative space, the model still largely detected the damaged corns. The YOLOv2 algorithm in Figure 5b included a false positive at the bottom left of the image in red. In Figure 5c the Faster R-CNN algorithm failed to detect two damage zones as indicated by the yellow bounding boxes with no number on the top of the region.

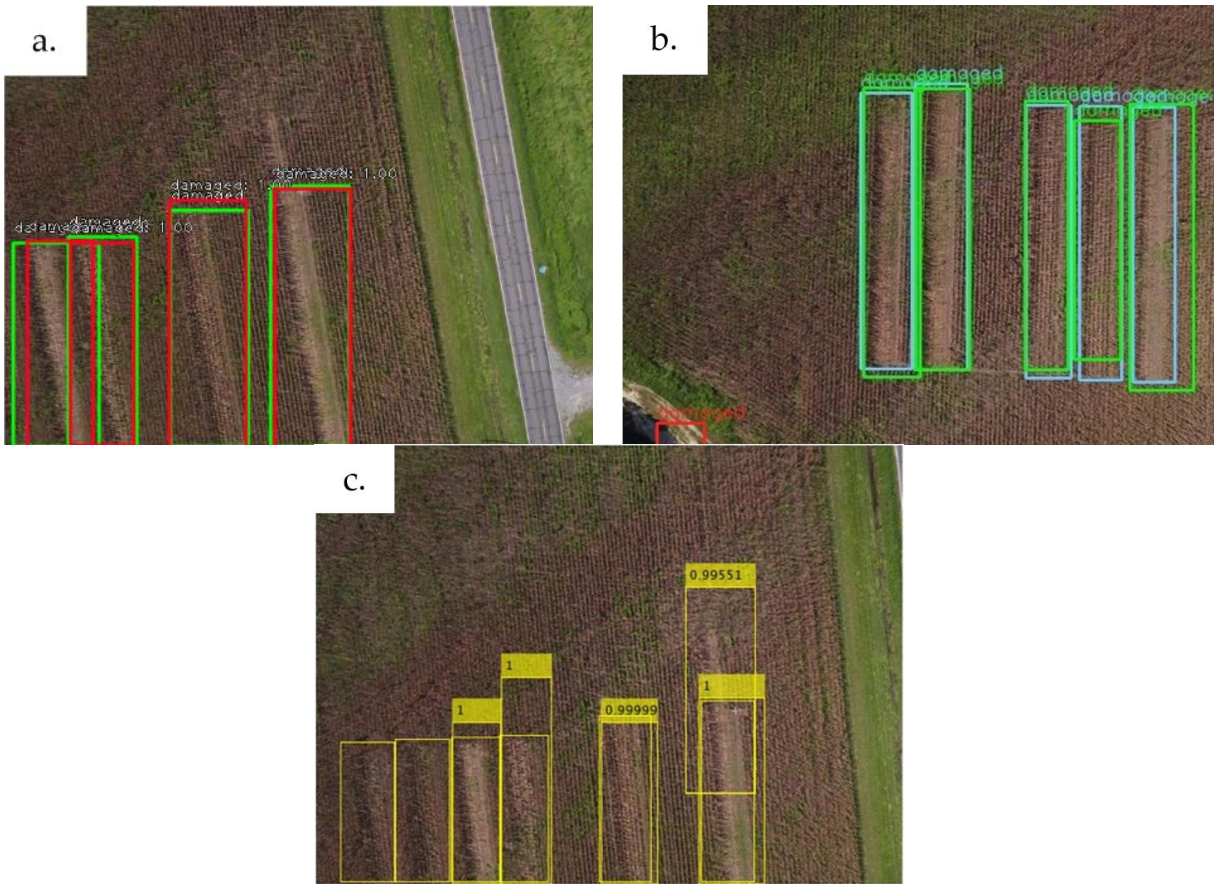

Figure 5. Ground truth vs. detection bounding boxes on an image from respective testing datasets; a. RetinaNet (red for ground truth, green for prediction and include confidence level and "damaged" tag for prediction boxes), b. YOLOv2 (blue for ground truth, green for prediction, red for false positive, and include "damaged" tag), c. Faster R-CNN (boxes without confidence level on the top for ground truth, boxes with confidence level on the top for prediction).

**AVERAGE PRECISION**

AP was used as an index to measures the overall capability of a model to detect predefined objects. The area under each precision-recall plot from different models were computed on test datasets to obtain AP (Table 4). Overall, Faster R-CNN showed lower precision at different growth stages compared to two other detectors. It was expected to have higher AP values for models created with a filtered dataset after eliminating turning images from training and testing datasets. Less preprocessing on incoming data was preferred so the algorithm can be trusted to robustly detect corn damage without additional data management. RetinaNet provided consistently high precision at different growth stages. Filtering out turning images did not show an improvement in detection accuracy. The reason is that models for not-filtered cases were provided by larger and more diverse datasets. Also, it can be seen RetinaNet network did not have difficulty to detect damaged areas even in turning images which is an extra benefit for using its models for this type of agricultural targets. However, filtering images before training models by YOLOv2 showed an improvement in model performance in most cases. The only inconsistency in the YOLOv2 models was for Model 1 which had higher detection power before data cleaning which could be caused by low number of images from samples on Stage 1. Since datasets were not very large especially in the first stages, the amount of precision can vary to some extent by changing the images in testing dataset. In most cases, model 5 had a relatively lower precision compared to the other models from the same group. The reason could be because of the high number of damaged areas with a small distance from each other which made it complicated for the detector to differentiate between background undamaged area and damaged areas. In addition, by time passing, weeds are emerging in the damaged regions created in earlier stages of the season (Figure 6).

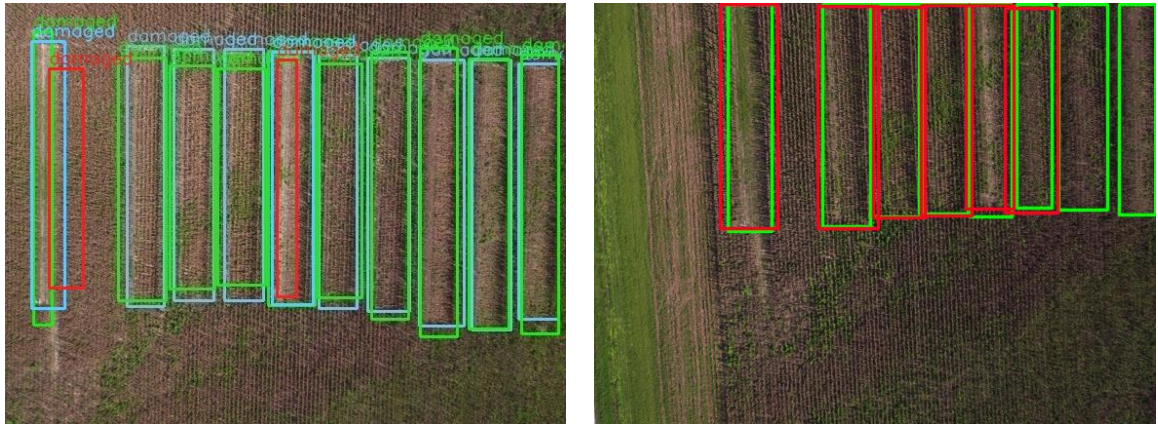

**Figure 6. Examples of errors in detection in late stages**

Overall, both YOLOv2 and RetinaNet were capable of accomplishing the task with promising results. It should be noticed that these AP values varies slightly after training. This difference can be more noticeable when training dataset is very small. Adding images with more variability can enhance the model predictive power.

**Table 4: Average precision on models from different object detectors at different stages (%)**

|  | Faster R-CNN |  | YOLOv2 |  | RetinaNet |  |
| --- | --- | --- | --- | --- | --- | --- |
|  | filtered | not-filtered | filtered | not-filtered | filtered | not-filtered |
| Model 1 | 66.04 | 53.82 | 90.91 | 97 | 82.97 | 98.43 |
| Model 2 | 64.83 | 65.38 | 85.37 | 73.76 | 93.87 | 92.63 |
| Model 3 | 77.29 | 73.56 | 95.5 | 89 | 90.2 | 81.39 |
| Model 4 | 64.99 | 72.08 | 91.28 | 83.49 | 79.26 | 83.97 |
| Model 5 | 43.98 | 14.47 | 90.14 | 55.99 | 73.24 | 82.37 |
| Model all | 52.81 | 65.93 | 89.9 | -* | 84.24 | 83.68 |

* YOLOv2 unable to successfully complete training without memory error.

## CONFUSION MATRIX

A confusion matrix describes the performance of a model on testing dataset for which the true values are known. Table 5 and Table 6 described the confusion matrix for the most precise models, RetinaNet and YOLOv2 at stage 1, in addition to more evaluation metrics calculated based on TP, FP, and FN.

**Table 5. Confusion matrix for the RetinaNet model at stage 1**

| | Positive | Negative |
|---|---|---|
| TRUE | 24 | N/A |
| FALSE | 5 | 0 |

| True Positive Rate | Relative False Positive Rate | Accuracy | F1 Score |
|---|---|---|---|
| 0.83 | 0.17 | 0.83 | 0.91 |

**Table 6. Confusion matrix for the YOLOv2 model at stage 1**

| | Positive | Negative |
|---|---|---|
| TRUE | 23 | N/A |
| FALSE | 6 | 0 |

| True Positive Rate | Relative False Positive Rate | Accuracy | F1 Score |
|---|---|---|---|
| 0.80 | 0.2 | 0.8 | 0.89 |

Based on the matrices for RetinaNet model and YOLOv2 at stage 1, there were no FN which means all damaged areas are detected and there is no missing object. There were 5 and 6 FP respectively, which is most likely because of the changes in ambient light which cause more complexity to differentiate the object from the background. By adding more images from new flights, detection error can be reduced. Also, since not-damaged areas cannot be quantified, TN are mentioned as N/A or "not applicable".

## DISCUSSION

Use little attention has been paid to cyberinformatic tools for monitoring corns near physiological maturity until a severe weather event destroys corn fields and prevents them from being harvested. New data-driven tools such as deep learning and computer vision can be deployed to skillfully identify the presence and severity of corn damage caused by severe weather. Resulting data can identify the areas of damage to be reported to farm owners, insurance companies, response agencies, or other relevant parties. This type of project can enhance the awareness and resiliency of growers at the time of natural disasters. Plus, corn damage can be detected in real-time or near real-time and provide meaningful information in no time or shortly after data collection. The expected end-point of this project is an on-line data processing tool in which a stitched georeferenced image of a field area with suspected corn damage is uploaded. The corn damage object detection model is applied to the uploaded image and a report is generated which describes the percent of the field which is damaged.

Zheng at al. (Zheng et al., 2018) compared different object detectors for identifying vegetable crops. Two of the tested object detectors were the same as those evaluated in this study: Faster R-CNN and YOLOv2. YOLOv2 performed better than Faster R-CNN, which is similar to the result of this study even though there were differences in conditions and imaging platform. (J. Zhou et al., 2016) utilized UAS imagery and vegetative indices from multispectral sensors to estimate hail damage in potato crops. Their results indicated that damage estimates were more accurate closer in time to the damage event, which is supported by the results presented here. Silleos et al. (Silleos et al., 2002) also applied vegetative indices to detect crop damage but used space-based sensors. The authors indicated that their method could identify fields which needed further inspection, but not discretely detect crop damage due to low resolution. The method described in this study identified specific sub-field regions which were damaged. Also, Kerkech et al. (2018) found UAV RGB images and deep learning approach useful for new method of detecting diseases in the vine-yard fields. The authors also mentioned data shortage a similar concern like this study which can be addressed in the future by collecting more samples and enriching image datasets. Chu, Starek, Brewer, Murray, et al. (2017) used UAS imagery and index-based technique and also structure-from-motion to determine the severity of the damage with R2 = 0.50. By comping this work and their work, it can be seen that computer vision and deep learning techniques can provide more accurate predictions on damage regions in 2D images. In addition, developing a model for damaged corn detection will result in better business application in real-time condition in future by using only 2D image without the necessity of 3D reconstruction and expensive processing.. Their effort was largely focused on small plot regions being used in phenotyping rather than large field areas targeted in this manuscript. Zhao et al. (2019) developed a model to assess rice lodging based on a deep learning UNet architecture and UAV imagery. The authors randomly cut one single image to many small samples to generate the training dataset while in the current study the data was collected at different stages during the growing season to consider the temporal effect in the training and testing datasets. The UAV In summary, the previous work by other researchers indicated the need for damage detection and described a gap in high resolution automated detection which can be filled with computer vision and deep learning techniques that can provide more accurate predictions.

One challenge identified across the object detectors under comparison was the relative decrease in performance as the corns advanced in growth stage, entered senescence, and became desiccated. The particular experimental design presented in this manuscript allowed natural and treatment variations to be captured. Visual observation of the corns indicate that at Stage 1 most of the field was green and the damaged regions were brown in color. At Stage 5 most of the field was brown along with the damaged regions. There was less variation in the image to allow damaged regions to be detected from non-damaged regions. This outcome pointed out that automated corn damage detection will perform best if damage occurs earlier in the growing season and indicated that data collection as close as possible to the damage event will improve damage detection accuracy. Beyond typical seasonal variation, weeds began growing in plots where damage were simulated which made segmentation more difficult as the non-target plants created challenges for the object detection models. This outcome was important to capture as this would happen under actual conditions and will need to be managed to increase damage detection accuracy.

## CONCLUSIONS

The main goal of this study was to test the feasibility of detecting damaged areas in the field remotely from a UAS using an affordable, easy-to-access RGB sensor. Computer vision tools and deep learning algorithms were assessed to identify corn damage automatically and remotely from UAS. In this study, three advanced object detectors, predeveloped for non-agricultural applications, were deployed. They were customized and trained to identify corn damage in images at different late-season growth stages. It was found that two object detectors, RetinaNet and YOLOv2, have robust capability for corn damage identification while Faster R-CNN exhibited decreased performance. Filtering images taken while the drone was turning showed improvement using YOLOv2 models. Minimizing data management processes prior to training is preferred and RetinaNet showed better performance without filtering. Modern UAS is a relatively new tool well-suited for quick monitoring the damage after a severe weather event. The awareness of growers and insurance agencies can be enhanced by providing a comprehensive report using remote sensing and object detection models. In future work, the current dataset will be extended by creating additional simulated data. A replicated, randomized complete block study with the same treatments presented previously will be established at two locations. The additional data should allow data segmentation as well as corn damage object detection. Different damage modes or severity should be detectable once a more diverse dataset is created.

## REFERENCES

Alsalam, B. H. Y., Morton, K., Campbell, D., & Gonzalez, F. (2017). *Autonomous UAV with vision based on-board decision making for remote sensing and precision agriculture.* Paper presented at the 2017 IEEE Aerospace Conference.

Amiri, H., Miller, T., & Savova, G. (2017). *Repeat before forgetting: Spaced repetition for efficient and effective training of neural networks.* Paper presented at the Proceedings of the 2017 Conference on Empirical Methods in Natural Language Processing.

Chu, T., Starek, M. J., Brewer, M. J., Masiane, T., & Murray, S. C. (2017). *UAS imaging for automated crop lodging detection: a case study over an experimental maize field.* Paper presented at the Autonomous Air and Ground Sensing Systems for Agricultural Optimization and Phenotyping II.

Chu, T., Starek, M. J., Brewer, M. J., Murray, S. C., & Pruter, L. S. (2017). Assessing Lodging Severity over an Experimental Maize (Zea mays L.) Field Using UAS Images. *Remote Sensing, 9*(9), 923.

Dauphin, Y., De Vries, H., & Bengio, Y. (2015). *Equilibrated adaptive learning rates for non-convex optimization.* Paper presented at the Advances in neural information processing systems.

Everingham, M., & Winn, J. (2011). The PASCAL visual object classes challenge 2012 (VOC2012) development kit. *Pattern Analysis, Statistical Modelling and Computational Learning, Tech. Rep*.

Falk, K. G., Jubery, T. Z., Mirnezami, S. V., Parmley, K. A., Sarkar, S., Singh, A., . . . Singh, A. K. (2020). Computer vision and machine learning enabled soybean root phenotyping pipeline. *Plant Methods, 16*(1), 5.

Hamidisepehr, A. (2018). *Classifying Soil Moisture Content Using Reflectance-based Remote Sensing.* (PhD Dissertation), University of Kentucky, Lexington, KY.

Hamidisepehr, A., & Sama, M. P. (2018a). *A low-cost method for collecting hyperspectral measurements from a small unmanned aircraft system.* Paper presented at the Autonomous Air and Ground Sensing Systems for Agricultural Optimization and Phenotyping III.

Hamidisepehr, A., & Sama, M. P. (2018b). Moisture Content Classification of Soil and Stalk Residue Samples from Spectral Data using Machine Learning. *Transactions of the ASABE, 62*(1), 1-8. doi:10.13031/trans.12744

Hamidisepehr, A., Sama, M. P., Turner, A. P., & Wendroth, O. O. (2017). A method for reflectance index wavelength selection from moisture-controlled soil and crop residue samples. *Transactions of the ASABE, 60*(5), 1479-1487. doi:https://doi.org/10.13031/trans.12172

Henderson, P., & Ferrari, V. (2016). *End-to-end training of object class detectors for mean average precision.* Paper presented at the Asian Conference on Computer Vision.

Hwu, W.-m., & Kirk, D. (2009). Programming massively parallel processors. *Special Edition, 92*, 157.

Jayas, D., Paliwal, J., & Visen, N. (2000). Review paper (AE—automation and emerging technologies): multi-layer neural networks for image analysis of agricultural products. *Journal of Agricultural Engineering Research, 77*(2), 119-128. doi:https://doi.org/10.1006/jaer.2000.0559

Kamilaris, A., & Prenafeta-Boldú, F. X. (2018). Deep learning in agriculture: A survey. *Computers and Electronics in Agriculture, 147*, 70-90. doi:https://doi.org/10.1016/j.compag.2018.02.016

Kazemi, V., & Sullivan, J. (2014). *One millisecond face alignment with an ensemble of regression trees.* Paper presented at the Proceedings of the IEEE conference on computer vision and pattern recognition.

Kerkech, M., Hafiane, A., & Canals, R. (2018). Deep leaning approach with colorimetric spaces and vegetation indices for vine diseases detection in UAV images. *Computers and Electronics in Agriculture, 155*, 237-243.

Koenig, K., Höfle, B., Hämmerle, M., Jarmer, T., Siegmann, B., & Lilienthal, H. (2015). Comparative classification analysis of post-harvest growth detection from terrestrial LiDAR point clouds in precision agriculture. *ISPRS journal of photogrammetry and remote sensing, 104*, 112-125. doi:https://doi.org/10.1016/j.isprsjprs.2015.03.003

Koirala, A., Walsh, K., Wang, Z., & McCarthy, C. (2019). Deep learning for real-time fruit detection and orchard fruit load estimation: Benchmarking of 'MangoYOLO'. *Precision Agriculture*, 1-29. doi:https://doi.org/10.1007/s11119-019-09642-0

Krizhevsky, A., Sutskever, I., & Hinton, G. E. (2012). *Imagenet classification with deep convolutional neural networks.* Paper presented at the Advances in neural information processing systems.

Li, H., Wu, Z., & Zhang, J. (2016). *Pedestrian detection based on deep learning model.* Paper presented at the 2016 9th International Congress on Image and Signal Processing, BioMedical Engineering and Informatics (CISP-BMEI).

Lin, T.-Y., Goyal, P., Girshick, R., He, K., & Dollár, P. (2017). *Focal loss for dense object detection.* Paper presented at the Proceedings of the IEEE international conference on computer vision.

Lu, H., Fu, X., Liu, C., Li, L.-g., He, Y.-x., & Li, N.-w. (2017). Cultivated land information extraction in UAV imagery based on deep convolutional neural network and transfer learning. *Journal of Mountain Science, 14*(4), 731-741. doi:https://doi.org/10.1007/s11629-016-3950-2

Ma, L., Shi, Y., Siemianowski, O., Yuan, B., Egner, T. K., Mirnezami, S. V., . . . Cademartiri, L. (2019). Hydrogel-based transparent soils for root phenotyping in vivo. *Proceedings of the National Academy of Sciences, 116*(22), 11063-11068. doi:https://doi.org/10.1073/pnas.1820334116

Mahajan, S., Das, A., & Sardana, H. K. (2015). Image acquisition techniques for assessment of legume quality. *Trends in Food Science & Technology, 42*(2), 116-133. doi:https://doi.org/10.1016/j.tifs.2015.01.001

Mohanty, S. P., Hughes, D. P., & Salathé, M. (2016). Using deep learning for image-based plant disease detection. *Frontiers in plant science, 7*, 1419. doi:https://doi.org/10.3389/fpls.2016.01419

Nolan, A., Park, S., Fuentes, S., Ryu, D., & Chung, H. (2015). *Automated detection and segmentation of vine rows using high resolution UAS imagery in a commercial vineyard.* Paper presented at the Proceedings of the 21st International Congress on Modelling and Simulation, Gold Coast, Australia.

Rahnemoonfar, M., & Sheppard, C. (2017). Deep count: fruit counting based on deep simulated learning. *Sensors, 17*(4), 905. doi:https://doi.org/10.3390/s17040905

Rançon, F., Bombrun, L., Keresztes, B., & Germain, C. (2019). Comparison of SIFT Encoded and Deep Learning Features for the Classification and Detection of Esca Disease in Bordeaux Vineyards. *Remote Sensing, 11*(1), 1. doi:https://doi.org/10.3390/rs11010001

Rebetez, J., Satizábal, H., Mota, M., Noll, D., Büchi, L., Wendling, M., . . . Burgos, S. (2016). *Augmenting a convolutional neural network with local histograms—a case study in crop classification from high-resolution UAV imagery.* Paper presented at the European Symp. on Artificial Neural Networks, Computational Intelligence and Machine Learning.

Redmon, J., & Farhadi, A. (2017). *YOLO9000: better, faster, stronger.* Paper presented at the Proceedings of the IEEE conference on computer vision and pattern recognition.

Redmon, J., & Farhadi, A. (2018). Yolov3: An incremental improvement. *arXiv preprint arXiv:1804.02767*.

Ren, S., He, K., Girshick, R., & Sun, J. (2015). *Faster r-cnn: Towards real-time object detection with region proposal networks.* Paper presented at the Advances in neural information processing systems.

Reyes, A. K., Caicedo, J. C., & Camargo, J. E. (2015). Fine-tuning Deep Convolutional Networks for Plant Recognition. *CLEF (Working Notes), 1391*.

Rhu, M., Gimelshein, N., Clemons, J., Zulfiqar, A., & Keckler, S. W. (2016). *vDNN: Virtualized deep neural networks for scalable, memory-efficient neural network design.* Paper presented at the The 49th Annual IEEE/ACM International Symposium on Microarchitecture.

Sammouda, R., Adgaba, N., Touir, A., & Al-Ghamdi, A. (2014). Agriculture satellite image segmentation using a modified artificial Hopfield neural network. *Computers in Human Behavior, 30*, 436-441. doi:https://doi.org/10.1016/j.chb.2013.06.025

Silleos, N., Perakis, K., & Petsanis, G. (2002). Assessment of crop damage using space remote sensing and GIS. *International Journal of Remote Sensing, 23*(3), 417-427. doi:https://doi.org/10.1080/01431160110040026

Stewart, S. R. (2017). *National Hurricane Center Tropical Cyclone Report*. Retrieved from https://www.nhc.noaa.gov/data/tcr/AL012017_Arlene.pdf

Stewart, S. R., & Berg, R. (2019). *National Hurricane Center Tropical Cyclone Report*. Retrieved from https://www.nhc.noaa.gov/data/tcr/AL062018_Florence.pdf

Tripicchio, P., Satler, M., Dabisias, G., Ruffaldi, E., & Avizzano, C. A. (2015). *Towards smart farming and sustainable agriculture with drones.* Paper presented at the 2015 International Conference on Intelligent Environments.

Vibhute, A., & Bodhe, S. (2012). Applications of image processing in agriculture: a survey. *International Journal of Computer Applications, 52*(2). doi:https://doi.org/10.5120/8176-1495

Zhao, X., Yuan, Y., Song, M., Ding, Y., Lin, F., Liang, D., & Zhang, D. (2019). Use of unmanned aerial vehicle imagery and deep learning unet to extract rice lodging. *Sensors, 19*(18), 3859.

Zheng, Y.-Y., Kong, J.-L., Jin, X.-B., Su, T.-L., Nie, M.-J., & Bai, Y.-T. (2018). *Real-Time Vegetables Recognition System based on Deep Learning Network for Agricultural Robots.* Paper presented at the 2018 Chinese Automation Congress (CAC).

Zhou, J., Pavek, M. J., Shelton, S. C., Holden, Z. J., & Sankaran, S. (2016). Aerial multispectral imaging for crop hail damage assessment in potato. *Computers and Electronics in Agriculture, 127*, 406-412. doi:https://doi.org/10.1016/j.compag.2016.06.019

Zhou, Y., Srinivasan, S., Mirnezami, S. V., Kusmec, A., Fu, Q., Attigala, L., . . . Schnable, P. S. (2019). Semiautomated feature extraction from rgb images for sorghum panicle architecture gwas. *Plant physiology, 179*(1), 24-37. doi:https://doi.org/10.1104/pp.18.00974